# 3D Programming of Patterned Heterogeneous Interface for 4D Smart Robotics

Kewei SONG[a], Chunfeng XIONG[a], Ze ZHANG[a], Kunlin WU[b], Weiyang WAN[b], Yifan WANG[b], Shinjiro UMEZU[a,c]*, Hirotaka SATO[b,*]

[a]*Graduate School of Creative Science and Engineering, Department of Modern Mechanical Engineering, Waseda University, 3-4-1 Okubo, Shinjuku-ku, Tokyo 169-8555, Japan.*

[b]*School of Mechanical and Aerospace Engineering, Nanyang Technological University, N3.2 – 01- 20, 65 Nanyang Drive, Singapore 637460, Singapore.*

[c]*Department of Modern Mechanical Engineering, Waseda University, 3-4-1 Okubo, Shinjuku-ku, Tokyo 169-8555, Japan.*

**\*Corresponding authors:**

*Shinjiro Umezu, Ph.D., Professor, E-mail: umeshin@waseda.jp*

*Hirotaka Sato, Ph.D., Professor, E-mail: hirosato@ntu.edu.sg*

# Abstract

Shape memory structures are playing an important role in many cutting-edge intelligent fields. However, the existing technologies can only realize 4D printing of a single polymer or metal, which limits practical applications. Here, we report a construction strategy for TSMP/M heterointerface, which uses $Pd^{2+}$-containing shape memory polymer (AP-SMR) to induce electroless plating reaction and relies on molecular dynamics, which has both shape memory properties and metal activity and information processing power. Through multi-material DLP 3D printing technology, the interface can be 3D selectively programmed on functional substrate parts of arbitrary shapes to become 4D electronic smart devices (Robotics). Microscopically, this type of interface appears as a composite structure with a nanometer-micrometer interface height, which is composed of a pure substrate layer (smart materials), an intermediate layer (a composite structure in which metal particles are embedded in a polymer cross-linked network) and a pure metal layer. The structure programmed by TSMP/M heterointerface exhibits both SMA characteristics and metal properties, thus having more intelligent functions (electroactive, electrothermal deformation, electronically controlled denaturation) and higher performance (selectivity of shape memory structures can be realized control, remote control, inline control and low voltage control). This is expected to provide a more flexible manufacturing process as platform technology for designing, manufacturing and applying smart devices with new concepts, and promote the development of cutting-edge industries such as smart robots and smart electronics.

## Introduction

Four-dimensional printing **(4D printing)** is considered a cutting-edge additive manufacturing **(AM)** technology for building customized smart devices with high integration and functional design [1-2]. It is showing good momentum in intelligent robots [3], innovative materials [4], wearable devices [5], intelligent electronics [6], biomedical engineering (artificial organs, artificial muscles, tracheal stents, drug delivery) [7], aerospace industry (self-expanding hinges, self-expanding trusses, self-expanding lens tubes, release devices) [8]. Among them, shape memory polymers **(SMPs)** open new horizons for smart structures, smart devices, and intelligent systems as promising smart materials with large deformability, excellent flexibility, and lightweight and high processability as intrinsic properties. Polymers with shape memory effect can be intelligently programmed to be controlled by specific physical fields such as thermal fields [9], light irradiation [10], electric fields [11], magnetic fields [12], chemical solvents [13] or other external stimuli [14] to recover the original shape and accomplish by this controlled process specific task. As the most widely used 4D printing material available, thermally responsive shape memory resin materials exhibit the advantages of a wide range of stimulation methods, simplicity of operation, and ease of implementation [15].

However, most heating strategies for thermally responsive 4D printing involve external heat sources. They cannot precisely control the heating properties to stably and activate specific 4D printing topologies to achieve actions selectively, thus hindering the application in real-world environments and limiting robots' intelligence. Some studies have attempted to achieve faster, more accurate, uniform heating strategies by building embedded circuits on the 4D printed part surface or directly. However, hand-constructing (applying conductive resin) patterns of heating elements or circuits embedded inside the part may affect the deformation mechanism in 4D printing. While stretchable electrodes constructed by conductive materials (by pleated nanoribbons/nanofilms [16] and by serpentine metal wires [17]) or controllable heating circuit material topologies (carbon nanotubes [18], graphene sheets [19], carbon lipids [20], conductive polymers [21], ionic gels [22] silver nanowires (NWs) [23]) still due to poor stretchability, poor material conductivity, conductive material processes and other deficiencies cannot meet the needs of manufacturing more intelligent devices thus not compatible with the more expansive 4D printing design space for smart robotics. The essence of these problems is the single function of the SMP interface, and activating its heat-driven shape memory properties requires additional functional topologies to assist in the realization. This makes the properties of the original substrate material and the additionally constructed active layer unable to be integrated, and macroscopically manifests as cracking and different mechanical properties. A new strategy [24-25] allows the constructing multifunctional interfaces with heterogeneous materials. The noble metal embedded in the active ink adsorbs the directionally deposited metal in the electroless plating reaction, resulting in the formation of a heterogeneous multi-material functional interface consisting of a polymer, an intermediate layer (the metal part is embedded in the polymer network), and a metal layer on its microscopic interface. Such an interface with both polymer and metal provides the possibility to realize the multifunctional intelligent effect in terms of building robots with completely new concepts.

This study reported a new three-dimensional **(3D)** selective construction strategy for thermoplastic SMPs/metal heterointerface **(TSMP/M heterointerface)** that transforms thermally actuated shape memory structures into electrically controlled ones. The prepared 4D active precursor achieved this, a multi-material digital light processing **(DLP)** 3D printing process **(MM-DLP3DP)** [25] coupled with electroless plating technology (**ELP**). The multi-material DLP 3D printing process allows the integrated fabrication of thermally driven SMPs and their configured active precursors according to the required topology. In the subsequent ELP, the active precursor topology exposed to the plating solution catalyzes and activates the metal deposition to trip the specific metal-clad plastic-metal composite structure. This method can construct a TSMP/M heterointerface pattern of any 3D shape **(3D programming)**, which has the functions of thermally driven deformation and metal wire electrodes so that the designed 4D electronics can replace the thermal field to complete the task. This multi-material 4D printing strategy 3D programmed on different substrate material structures using TSMP/M heterointerface (pattern) as a whole is expected to break through the design limitations of traditional intelligent systems and endow them with broader possibilities. Its strategic universality is also expected to promote new concept device technology as platform technology expands to construct other intelligent interfaces for intelligent robots.

## Selective 3D programming of TSMP/M heterointerface

The principle of the construction of the proposed 4D electronic structure (composite structure composed of SMPs and metal patterns) is achieved by controlling the surface activity of different SMP topologies (catalyzing the activity of electroless metal deposition). To achieve this, we first prepared new ultraviolet photosensitive shape memory resin **(UV-SMR)** and further obtained active precursors with shape memory resins **(AP-SMR)**. Such materials can be constructed to activate electroless metal deposition to form metal-clad topologies while possessing shape-memory properties. TSMP/M heterointerface is expected to combine the properties of functional polymers and metals to achieve more intelligent functions. With MM-DLP3DP, more possibilities can be added to 4D printing by implementing the 3D programming of the TSMP/M heterointerface. Smart polymers with functional structures and metal topologies with electronic properties will enable the intelligent transition of existing 4D printed structures into 4D electronic devices. **Figure 1** shows the schematic diagram of the proposed concept of 3D selective construction of TSMP/M heterointerface (3D programming of substrate parts using TSMP/M heterointerface) and 4D electronics using the MM-DLP3DP process. As shown in **Fig. 1 a**), the MM-DLP3DP process can selectively construct composite structures composed of UV-SMR and AP-SMR (**Figure 1 e) upper part**). In the subsequent ELP, AP-SMR can be selectively covered with metal to form a specific pattern of TSMP/M heterointerface (**Figure 1 e) middle part**). The 3D topology composed of TSMP/M heterointerface can realize the transition from thermal field drive to electric field drive and become 4D electrons. The concept is shown in **Fig. 1 b)**, the middle part of the Eiffel Tower model is 3D programmed with TSMP/M heterointerface so that the shape can be restored under the control of the current after being shape-programmed. The patterning of the TSMP/M heterointerface is expected to activate the potential properties of 4D materials to achieve precise control of their deformation process. The output of its driving power can be expressed as an electric heating formula, **Equal. 1)**:

$$Q = U^2 t / R \qquad\qquad 1)$$

Where, U is the voltage value driving the TSMP/M heterointerface topology; t is the power-on time; R is the total resistance value of the metal part of the heterointerface topology; Q is the total calorific value of the driving time.

Since the metal coating obtained using ELP has good uniformity, resistivity and ductility [26], TSMP/M heterointerface is of great significance for realizing low-voltage control, precise control and off-line control of 4D devices. As the substrate of the TSMP/M heterointerface, synthesizing SMPs is an important basis. Compared with the existing research [27], the materials in this technical scheme need to have certain activity based on having positive functions and properties (Smart active materials). Unlike other technical routes that utilize simple polymer network synthesis, our strategy for preparing thermally SMPs uses semi-crystalline polymers [poly(caprolactone), PCL] as molecular switching phase. By adding PCL semi-elasticity solution to the polymer network of ABS-like UV resin body (thermoplastic polymer network), we have obtained UV-SMR, which can be cured and molded under the irradiation of UV light with a wavelength of 405 nm. In this new SMPs system, PCL as reversible phase for fixing the temporary shape has been proven to have positive performance [28]. The key step for successfully introducing PCL into the photopolymer network is to use chloroform as a solvent to realize the complete and adjustable dissolution of PCL to obtain a PCL solution.

As a stationary phase for holding the permanent shape, ABS-like resin comprises a comprehensive composite polymer network with high mechanical properties (**Table S1**) [29]. As shown in **Fig. 1 c)**, as a description of the newly synthesized SMPs system, the functional polymer network composed of photo initiator, monomer, polymer (all three are provided by ABS-like resin), and PCL. After polymerization under UV, PCL was embedded in the cross-linked network as a functional phase (PCL crystal) to realize the effect of shape memory. By adding a certain amount of $Pd^{2+}$ based activation solution ($Pd^{2+}$-AS), active precursors with both shape memory properties and catalytic ELP could be obtained. The proposed TSMP/M heterointerface can be built on the functional substrate as a whole cooperated with MM-DLP3DP and coupled to form a new functional interface to realize the designed smart function. **Figure 1 d)** shows the construction process of the TSMP/M heterointerface in UV-SMR (or other smart material substrates). The fabricated UV-SMR and AP-SMR of MM-DLP3DP exhibit a composite structure at their interface. In the ELP process, AP-SMR gradually realizes the deposition of metal particles and finally forms the TSMP/M heterointerface.

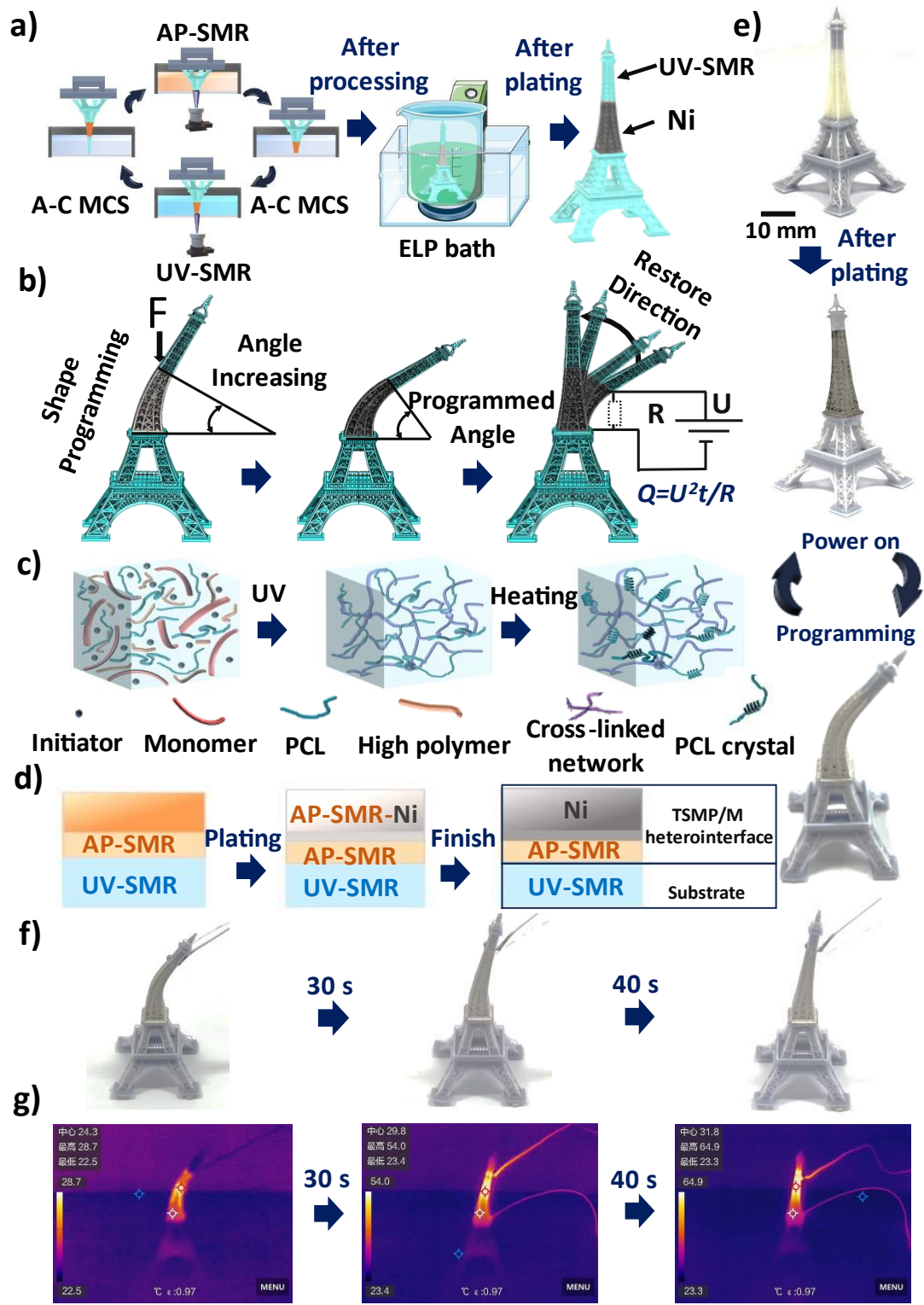

**Figure 1. The schematic diagram of the proposed concept of 3D selective construction of the TSMP/M heterointerface and 4D electronics using multi-material 3D printing.** a) The MM-DLP3DP process can realize the fabrication of UV-SMR/AP-SMR composite structures with arbitrary complex shapes, which can be selectively deposited metal in subsequent ELP to build 3D TSMP/M heterointerface; b) The concept of 4D electronics realizes the function by selectively constructing

the TSMP/M heterointerface on the functional substrate structure to control the electrical activity of specific topology; c) The newly prepared UV-SMR system is mixed with photo initiator, monomer, polymer and PCL, in which PCL is embedded in cross-linked network after UV exposure to achieve shape memory effect (when heated, PCL work as PCL crystal); d) The construction process (Under the macro perspective) of the TSMP/M heterointerface on functional substrate (UV-SMR). MM-DLP3DP can build the TSMP/M heterointerface on any photosensitive polymer functional material using AP-SMR; e) UV-SMR/AP-SMR composite structure, composite structure with TSMP/M heterointerface and the Eiffel Tower part after selective shape programming; f) Electronically controlled restoration of a model of the Eiffel Tower programmed with selective shape memory to its original shape; g) The topology with TSMP/M heterointerface behaves as a controllable thermal field in the energized state.

On the whole, the significance of selective construction lies in more precise control of the target structure, in which the electroactive topological part has better controllability. After the Eiffel Tower part with TSMP/M heterointerface is powered on, the PCL crystal in its TSMP/M heterointerface will be selectively activated. It thus can perform shape programming **(Figure 1 e) lower part)**. After power on again, the part wrapped by metal (total resistance 7 Ω) will selectively heat up (driving voltage 1.5 V) to drive the programmed topology to restore its original shape. **Figure 1 f)** shows the process (**Video 4**), the shape recovery process only takes about 70 s, and the heating field during the whole process is precisely controlled within the target range which was confirmed by thermal imaging equipment (**Figure 1 g)**).

As the basis of the TSMP/M heterointerface construction, the performance of UV-SMR determines its structural properties (mainly mechanical properties) and functionality. This method for preparing 4D functional polymers is simple, and the configured PCL solution, UV-SMR and AP-SMR all exhibit positive stability (**Figure S1**) and viscosity (**Figure S2**), which is critical for long-term DLP printing [30]. The addition of PCL solution and $Pd^{2+}$ solution did not affect their original absorbance properties. Although the absorbance peaks of UV-SMR and AP-SMR with different compositions changed in the fourier transform infrared (FTIR) device (**Figure 2 a)**), the peak positions did not change significantly, predicting an effective DLP 3D printing molding capability 405 nm UV radiation. **Figure 2 b)** shows a model of the Eiffel Tower manufactured by UV-SMR and its process of achieving shape memory (**Video 1**), with the skeletonized structure of the part intact. The favorable printing accuracy (**Figure 2 c)**) of the material allows complex thin-walled parts (**Figure 2 d)**) to be successfully manufactured and deformation (**Video 2**). This accuracy is not significantly affected by the addition of $Pd^{2+}$ solution (**Figure 2e)**), and microfabrication can also be achieved by AP-SMR (**Figure 2 f)**) (**Video 3**). These will support the basic mechanical properties of the TSMP/M heterointerface and provide a solid material foundation for its extensive construction design.

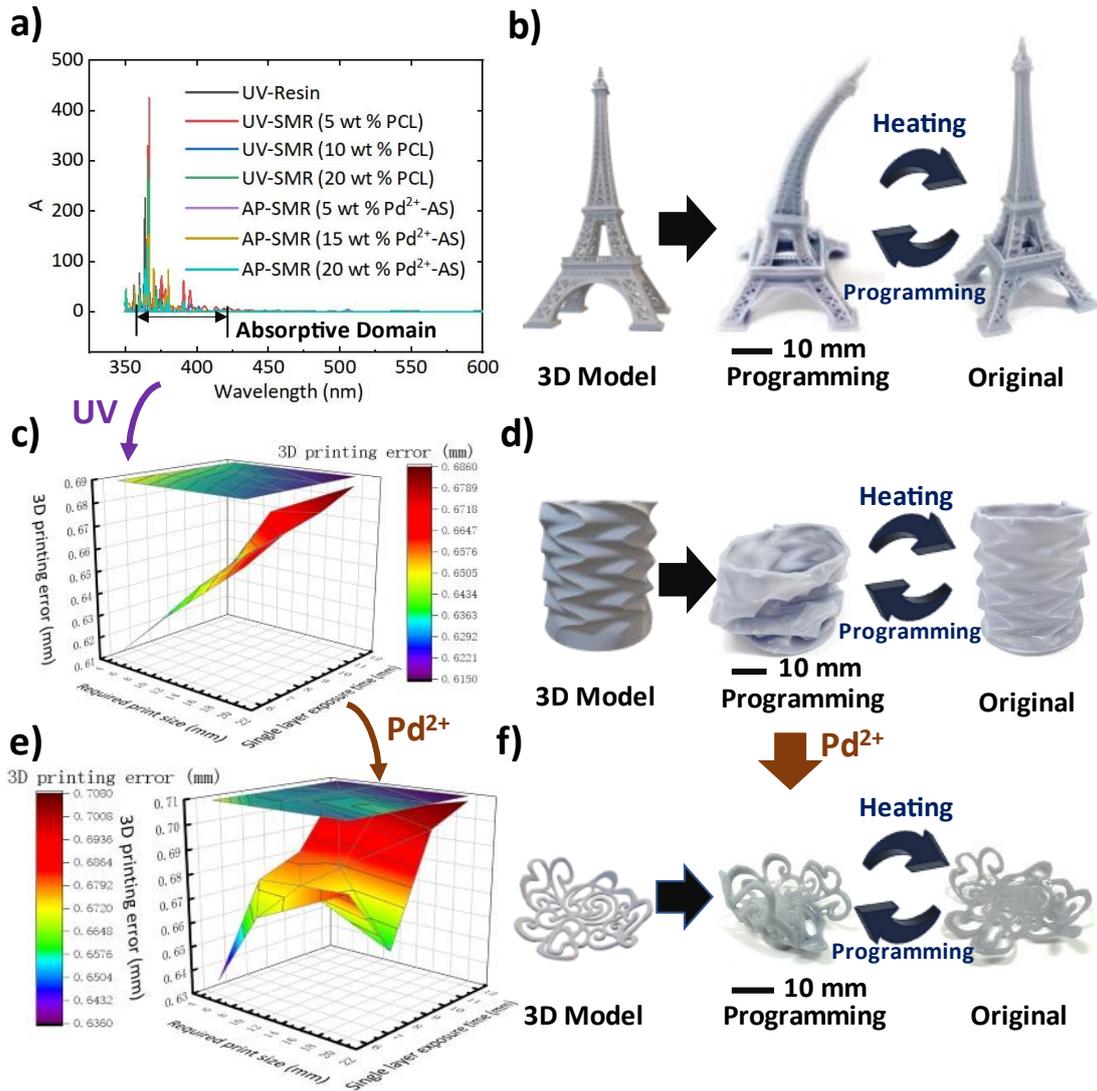

**Figure 2. Formability of UV-SMR and AP-SMR.** a) FTIR absorption curves of UV-Resin, UV-SMR and AP-SMR in the low wavelength range. The addition of PCL solution and Pd²⁺ solution does not change the peak position of the polymer; b) The printing effect of the classic Eiffel Tower model composed of UV-SMR, which retains the hollow structure; c) The printing accuracy of UV-SMR is basically kept within 6% error range; d) The printing effect of thin-walled parts composed of UV-SMR, which supports aggressive molding properties; e) The printing accuracy of AP-SMR is basically kept within 7% error range; f) The printing effect of microstructure parts composed of AP-SMR.

The addition of PCL solution changed the Young's modulus [31,32] of the original ABS polymer (**Figure 3 a)**), and the influence is most significant when its concentration reaches 20 wt%. In contrast, the Young's modulus is less than 25 MPa to below 5 MPa. This means the formed UV-SMR structure is softer but more easily deformed, while we chosed 15 wt% as the threshold for configuring UV-SMR. Nevertheless, the effect of Pd²⁺ active solution (Pd²⁺-AS) was carried out when the concentration of PCL was 20 wt%, since the influence of one PCL solution as a variable was stable. As shown in **Fig. 3 b)**, the effect of Pd²⁺-AS on the Young's modulus of AP-SMR is insignificant.

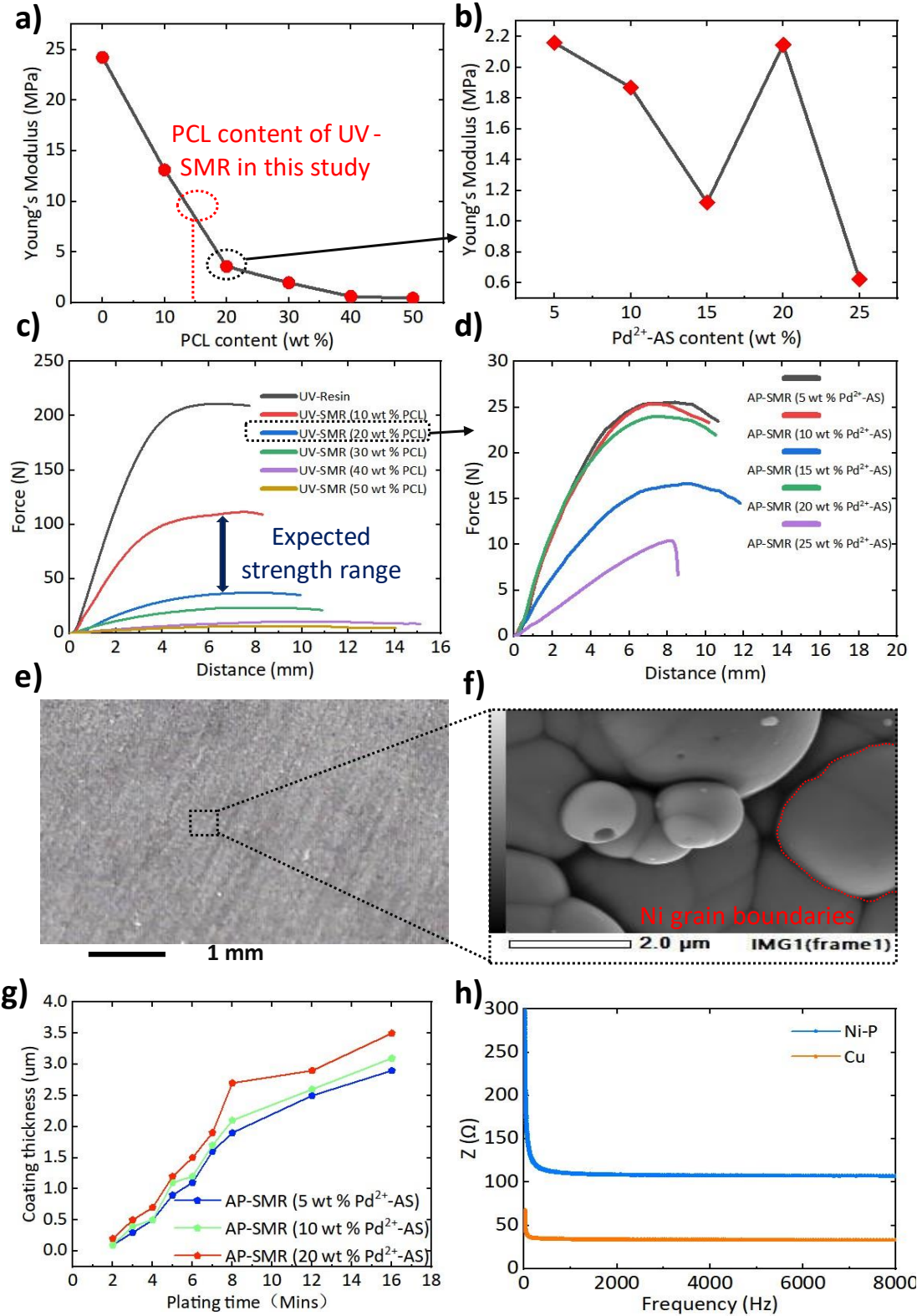

**Figure 3. The TSMP/M heterointerface, UV-SMR, AP-SMR and properties of metal layers.** a) Effect of PCL solution concentration on Young's modulus of UV-SMR; b) Effect of Pd$^{2+}$-AS concentration on Young's modulus of AP-SMR when the content of PCL solution in UV-SMR is 20 wt% ; c) The effect of PCL solution concentration on the tensile strength of UV-SMR; d) When the PCL solution content in UV-SMR is 20 wt%, the effect of the concentration of Pd$^{2+}$-AS on the tensile strength of AP-SMR ; e) The microstructure of the metal coating (Ni), which is macroscopically represented as a dense metal film stacked by Ni lattices; f) The element distribution of the metal coating (Ni); g) ELP time for different concentrations of $Pd^{2+}$-AP. The effect of the growth rate of the coating film; h) The impedance characteristic curve of the Ni and Cu coating obtained.

Since the PCL solution was obtained using chloroform, the intensity of UV-SMR was reduced compared to that of the original ABS-like resin. As described in **Fig. 3 c)**, as the content of PCL increases, the tensile strength decreases from the original 46 MPa to 8.67 MPa at 20% since chloroform will destroy the cross-linked network to a certain extent, but acceptable. In the bending test results (**Figure 3 d)**), the decrease in strength is not significant, because the introduction of PCL increases the toughness of the whole UV-SMR structure to a certain extent.

After the ELP process was completed, the surface of the AP-SMR will be deposited a layer of metal coating formed by stacking crystals to form the metal layer of the TSMP/M heterointerface. **Figure 3 e)** shows the morphology of the metal film after ELP of Ni, in which multiple Ni overlap and accumulate into a uniform (macroscopically) metal coating. The metal coating was dense, free of pores and cracks, and the curved coating was also free of cracks and peeling **(Figure 3 f))**. The metal layer was made up of Ni crystals and the thickness of this film is determined by the ELP time, and the increase of $Pd^{2+}$ concentration will increase the film deposition rate (speed), like shown in **Fig.3 g)** but has little effect on the appearance (**Figure S9**). **Figure 3 h)** shows the impedance characteristic curves when the metal layers are Ni and Cu, and their resistivities are 11.07 μΩ·cm and 4.89 μΩ·cm, respectively, which are sufficient to function as electrons. When the metal deposition is completed, the area is programmed to become the TSMP/M heterointerface, which is electrically active. Building metal layers directly on functional insulators promises to lead to smart structures with better electrical properties, whether as functional electrodes (Compared to conducting polymer strategies [38]) or electronic media for information processing. This will promote the practical industrial application of 4D printed parts as smart devices, especially in microelectronics, embedded electronics, and low-power electronics.

## Construction mechanism and structure of micro-nano TSMP/M heterointerface

The construction of the TSMP/M heterointerface is based on the $Pd^{2+}$ embedded in the polymer cross-linked network in the microscopic state of the active precursor, the process is described in **Fig. 4 a)**. The $Pd^{2+}$-AS needed to configure the active precursor is formed by dissolving $PdCl_2$ in saturated $NH_4Cl$ solution. Partial Pd (II) cations have a square inverse prism geometry defined by four $Cl^-$, then form $[PdCl_4]^{2-}$ ligands in active solution, and this reaction is generally described as **Equal. 2)**. Due to the high electrode potential, $[PdCl_4]^{2-}$ ligand cannot act as a catalyst.

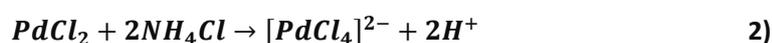

$$PdCl_2 + 2NH_4Cl \rightarrow [PdCl_4]^{2-} + 2H^+ \qquad \textbf{2)}$$

**Figure 4. b)** reflects the X-ray diffraction **(XRD)** of different polymers after printing. The AP-SMR with different concentrations of Pd²⁺-AS has a peak at 2θ = 32.62, which confirms the existence of Pd (Ⅱ). In order to obtain Pd (0) with higher catalytic activity, different from our previous research [25], we performed a reduction operation on it before performing ELP, using $NaH_2PO_2$. The solution has the process expressed by the **Equal. 3)** [33].

$$Pd(Ⅱ) + H_2PO_2^- + H_2O \rightarrow Pd(0) + H_2PO_3^- + 2H^+ \qquad 3)$$

This process is marked as process (0) in **Fig. 4 a)** as a pre-processing stage. **Figure 4 c)** show the X-ray photoelectron spectroscopy **(XPS)** curves of the printed AP-SMR parts before and after processing (chemical reduction). **Figure 4 c) (left side)** shows the primary XPS region of O1s before chemical reduction, showing the binding energies range of eV = 532.4-537.7, with the peak near eV = 535 shows the presence of carbon-oxygen bonds, carbon-oxygen double bonds and water or chemisorbed oxygen species [34]. After chemical reduction (pretreatment), the peak position of its XPS spectrum did not change significantly, and the Binding Energy (eV) increased. However, it was not found that the functional groups of AP-SMR participated in the reaction. **Figure 4 d)** shows the XPS curves of Pd element in AP-SMR of different components before and after chemical reduction. The binding energy peak group changes from 340 eV (Pd (0)), 343 eV (Pd (Ⅱ)) and 344 eV (Pd (Ⅱ and Pd (Ⅳ))-dominated to 335 eV (Pd (0)), 338 eV (Pd (Ⅱ)) and 344 eV (Pd (Ⅱ) and Pd (Ⅳ))-dominated. Pd (Ⅳ) may be due to contamination or the addition of chloroform. Although the peak with low concentration of Pd²⁺-AS (5 wt%, 10 wt%) is not obvious, it supports the fact that Pd (Ⅱ) embedded on the AP-SMR surface is reduced to Pd (0) by sodium hypo phyllite partly. The reduction is incomplete and depends on the concentration, temperature and soaking time of sodium hypothionite. However, it reveals that the sodium hypothionate solution can be immersed in the cross-linked network of a certain depth on the surface of AP-SMR to reduce Pd (Ⅱ). This provides a potential way to obtain AP-SMR with higher catalytic activity, so that the TSMP/M heterointerface can be constructed more efficiently. Compared to Pd (Ⅱ), Pd (0) has empty and active d-orbitals. During its induction of catalytic electroless (Ni) plating, Pd (0) activity should be one of the most important factors determining the extent of $H_2PO^{2-}$ interaction with its surface. In the case of Pd (0) embedded in the cross-linked network of functional polymers, the plating solution penetrates the cavities in the cross-linked network, its Hs-orbitals interact with d and p-orbitals, and Ps-orbitals interact with Op-Orbit. This interaction creates an effective antibonding structure and a strong P-H cleavage promotion in the P-H bond. This change will promote the distortion of the $H_2PO^{2-}$ planar structure. The distortion effect originates from the coordination of the lone pair of electrons in OH⁻ to the empty p-orbital of P, which will lead to the construction of a new P-O bond. $H_2PO_2^-$ spontaneously oxidizes to form $H_2PO^{3-}$ dehydrogenation and charge transfer. The charge transfer step occurs between $H_2PO^{3-}$ and Ni (Ⅱ) charge transfer, that is, Ni (Ⅱ) gets electrons and is reduced to Ni atoms. Its construction process can be described as **Equal. 4)**, but it contains four processes (process (1), process (2), process (3), process (4) corresponds to **Equal. S2)-S5)** respectively.).

$$Ni^{2+} + H_2PO_2^- + H_2O \xrightarrow{Pd} 3H^+ + H_2PO_3^{2-} + Ni \qquad 4)$$

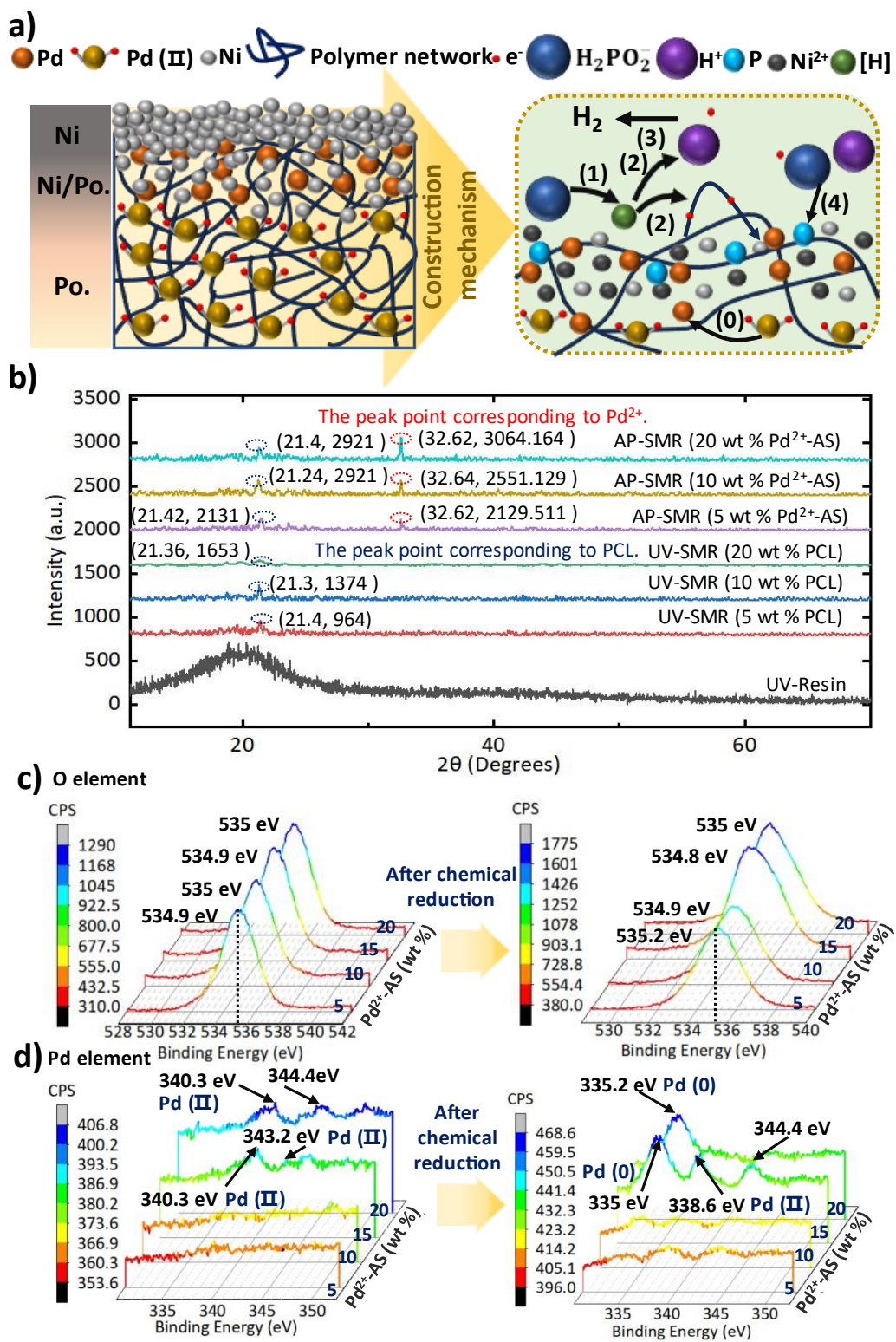

**Figure 4.** **The mechanism of AP-SMR inducing electroless plating and inducing the construction of the TSMP/M heterointerface in this study.** a) The chemical reaction process during the construction of the TSMP/M heterointerface, Po. means polymer; b) XRD curves of UV-SMR and AP-SMR of different components (containing different concentrations of PCL) and AP-SMR (containing different concentrations of $Pd^{2+}$-AP); c), d) XPS curves of AP-SMR parts with different concentrations before and after chemical reduction reaction (oxygen element and Pd element).

The intelligence of the TSMP/M heterointerface is reflected in its composite cross-section, which simultaneously has the electronic properties of metals and the functionality of SMPs. The cornerstone of its intelligent function as a whole is the adhesion between the metal layer and the functional polymer layer in its interface. Based on the improvement of the process of building 3D metal wires on normal resin, in order to make AP-SMR exhibit higher catalytic activity and improve adhesion, we pre-treated the parts before ELP. The pretreatment process consists of chemical etching for surface roughening to expose more $Pd^{2+}$ and chemical reduction process for reducing $Pd^{2+}$ to Pd simple substance. They were completed by 20 wt% sulfuric acid and 40 wt% $NaPO_2H_2$ solutions, respectively, to directly dip the parts constructed by AP-SMR into them. It is worth noting that there is no need to pay attention to selectivity in the operation process, and its selective activation is determined by the original selectivity of $Pd^{2+}$ firmly embedded in AS-SMR, so the selectivity of metal construction will not be affected by the pretreatment process. Macroscopically, the TSMP/M heterointerface after pretreatment shows the improvement of the adhesion between the metal layer and the AP-SMR layer. **Figure 5 a)-c)** show the adhesion of the metal plating (Ni) in the TSMP/M heterointerface under the plating peel test method (D2095-96 standard). Please refer to **Table S6** for testing and experimental methods.

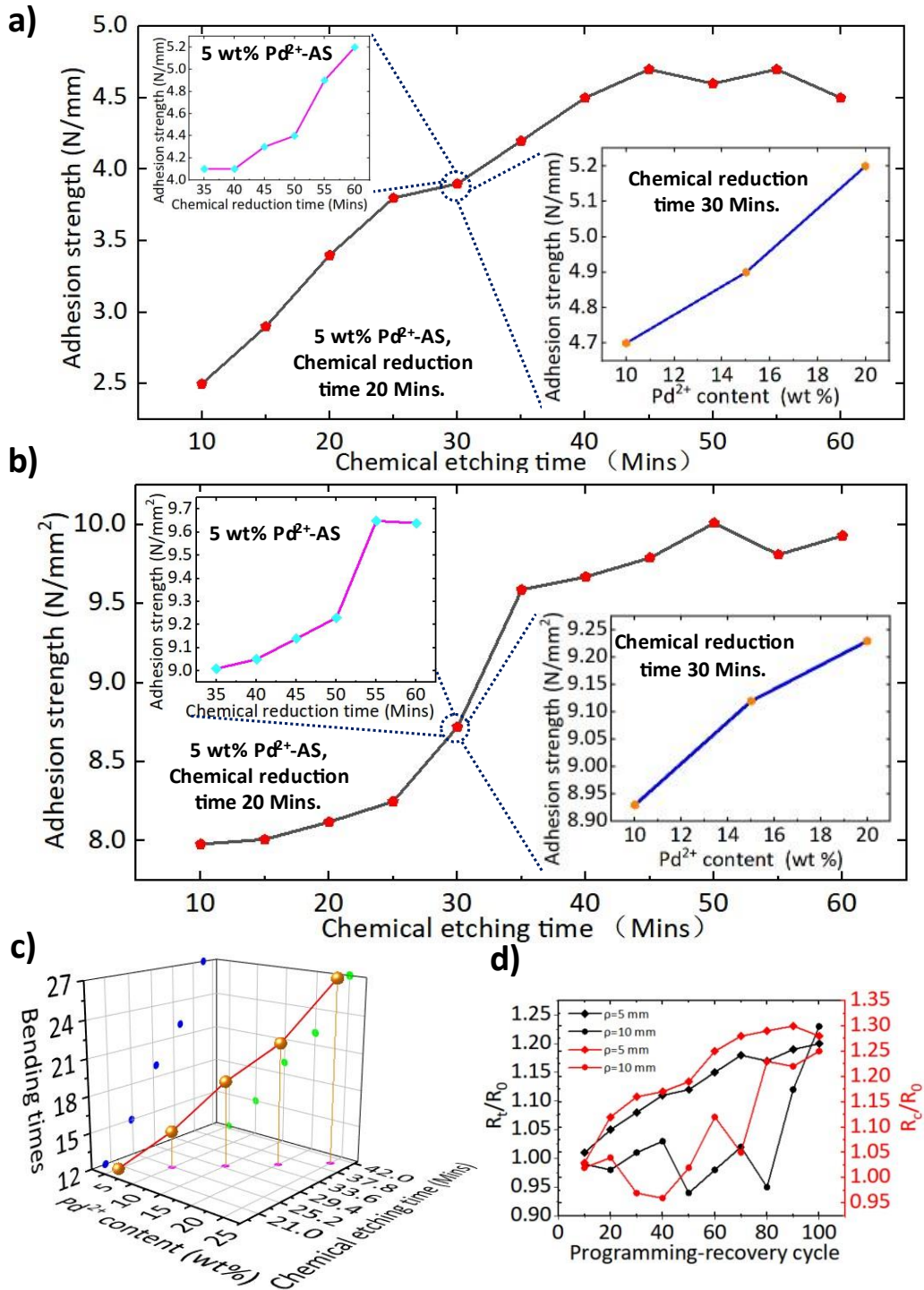

**Figure 5. Adhesion test results of metal layer and AP-SMR layer in the TSMP/M heterointerface and its relationship with chemical etching time, chemical reduction time and Pd²⁺ concentration.**
a) Under the plating peel test method, the test results of the adhesion force; b) Under the vertical peel test method, the test results of the adhesion force; c) Under the bending peel test method, the test results of the adhesion force. Under the three adhesion evaluation mechanisms, they all show positive correlations with chemical etching time, chemical reduction time and Pd²⁺ concentration; d) Flexible electronics fatigue test results, $R_0$ is the initial resistance value before the bending test; $R_t$ is the resistance value in the tensile state; $R_c$ is the resistance value in the compression state.

The most intuitive factor is the chemical etching time. By immersing in sulfuric acid, the surface of AP-SMR will be corroded and its microscopic surface roughness will increase, which will help the adhesion of the metal coating. Nevertheless, when the chemical etching time increases to about 45 Mins (5 wt% Pd$^{2+}$-AS, Chemical reduction time 20 Mins.), the improvement effect of its adhesion force decreases, indicating that excessive etching cannot further improve its adhesion force, as shown in **Fig. 5 a)**. And when we fixed the time of chemical etching and reduction (chemical etching time is 30 Mins, chemical reduction time is 30 Mins), the Pd$^{2+}$ content in AP-SMR is gradually increased, and its adhesion shows a positive correlation trend. Similarly, the improvement of the adhesion is positively correlated with the chemical reduction time (chemical etching time is 30 Mins, 5 wt% Pd$^{2+}$-AS). Similar experimental results were confirmed again in the vertical peel test **(Figure 5 b))**, and the adhesion showed an increase relationship with the increase of Pd$^{2+}$ concentration, chemical etching time and chemical reduction time. As far as the TSMP/M heterointerface is concerned, its electrical activity process relies on the continuous programming of the entire structure and the shape recovery after stimulation, which is a process of undergoing continuous deformation. Rather, its dynamic adhesion (the ability to remain adhered after multiple deformations) is more indicative of its performance, which can be evaluated by Bending peel test. **Figure 5 c)** shows the number of times the TSMP/M heterointerface is repeatedly bent 90° without bubbles in the metal layer, which is also affected by the chemical etching time, chemical reduction time and Pd$^{2+}$ concentration. Among the above three influencing factors, the chemical etching time can improve the adhesion between the metal and the functional polymer by changing the surface roughness of AP-SMR, while the chemical reduction time and the concentration of Pd$^{2+}$ are by affecting the AP-SMR. Activation of the SMR surface achieves an increase in its adhesion, which is very important to the concept of 4D electronics. The topology programmed by TSMP/M heterointerface becomes 4D electrons, which can realize the shape programming-shape recovery process driven by current, which is actually the process of continuous bending and deformation of flexible electrons. Excessive use of 4D electrons may cause the metal layer in the heterogeneous interface to separate from the plastic, which is manifested as a decrease in conductivity. To assess the service life of 4D electronics, Flexible electronics life test is carried out (**The detailed experimental methods are shown in Table S8**) to measures the resistance increase rate after bending the TSMP/M heterointerface for a certain number of times. As the **Figure 5 d)** shows, regardless of the curvature of $\rho = 5\,mm$ or $\rho = 10\,mm$, the resistance change remains within the acceptable change rate range after up to 100 bending tests.

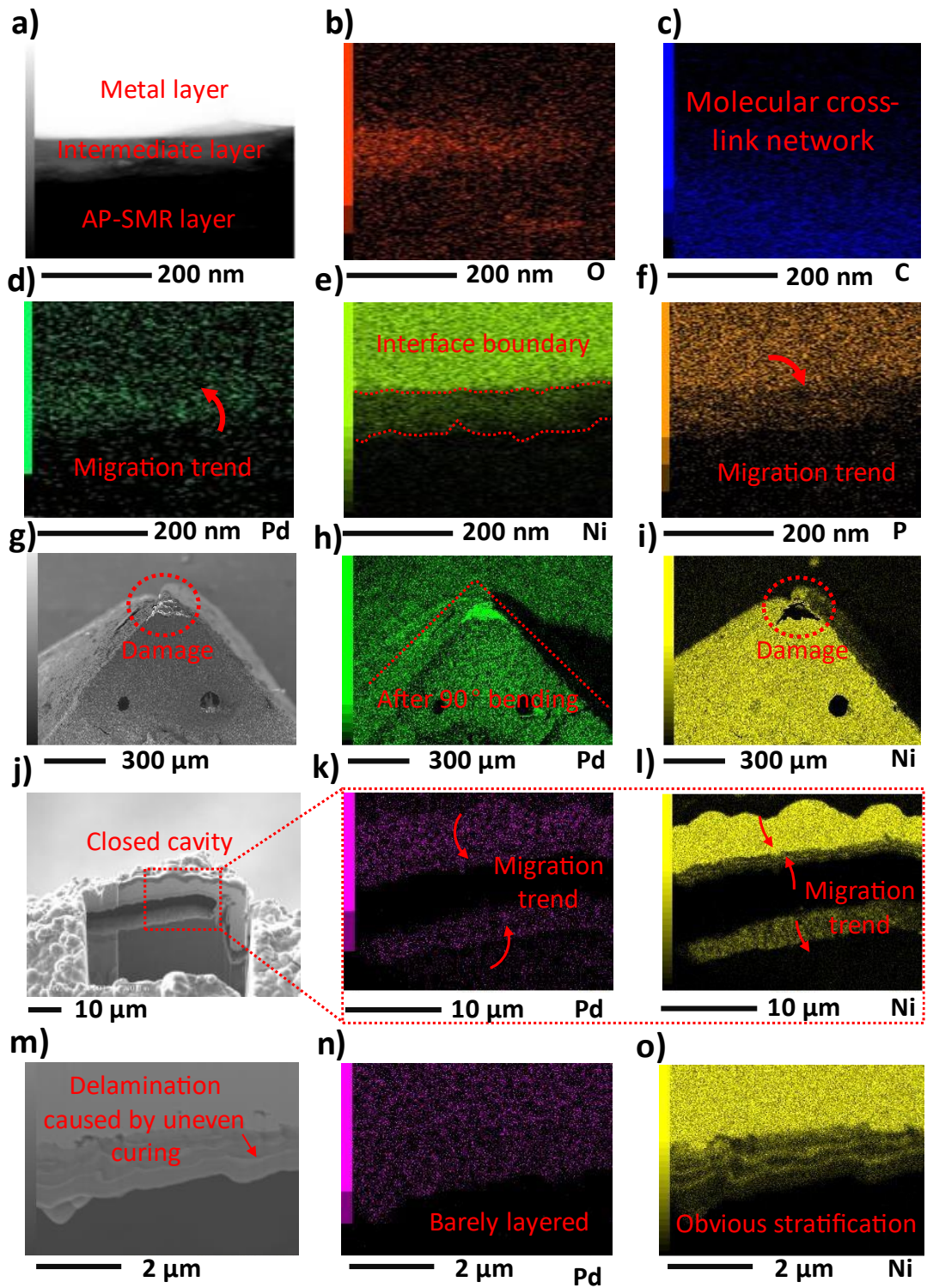

**Figure 6. The composite section microstructure of the TSMP/M heterointerface characterized by EDS.** a) Schematic diagram of the interface (cross-section) structure under a high-power electron microscope, which is composed of a metal layer, an intermediate layer and an AP-SMR layer; b) The division of O element; c) The division of C element; O element and C element together characterize the microscopic cross-sectional structure of the crosslinking network; d) The division of Pd element, showing the tendency of outward transition; e) Ni element The division of Ni embedded in AP-SMR to form an intermediate layer and shows a deep trend; f) the division of P element, supporting the transition trend of Ni; g) the microstructure of the part after folding 90°, in which a small amount of Ni There is damage; h) The division of Pd element after the part is folded at 90°, confirming that there is no damage in AP-SMR; i) The division of Ni element after the part is folded at 90°; j) The internal cavity structure formed due to the molding defect ; K) Partition of Pd element in the cavity structure; l) Partition of Ni element in the cavity structure; m) Layered structure due to uneven photocuring; n) Partition of Pd element in the layered structure , there is no stratification; o) The division of the Ni element in the stratified structure shows a well-defined and regular stratification phenomenon.

During the Pd-catalyzed electroless plating reaction, the composite section microstructure of the TSMP/M heterointerface is gradually formed. **Figure 6 a)** shows its cross-sectional structure under high magnification, which exhibits a remarkable three layers. Different from the well-defined metal layer and AP-SMR layer in its macrostructure, there is a chaotic intermediate layer between them, in which Ni particles are embedded in the AS-SMR layer with a deep tendency (**Figure 6 b)-f)**). This phenomenon was preliminarily discussed in the study of plastic-metal composite structures [25]. For the TSMP/M heterointerface, after PCL is mixed into the cross-linked network, its structural density will increase, and the cross-linked network will be filled and lose part of its water absorption properties. The formation of the intermediate layer is attributed to the adsorption of Ni particles produced by Pd-catalyzed electroless electrolysis. This adsorption can be achieved through the chemical bond between Pd and Ni or the intermetallic interaction, where the attractive force of Pd on the Ni surface is mediated through the interaction force. Under this interaction force, Ni transitions with a tendency to embed into the cross-linked network, while Pd transitions with a tendency to be sucked out of the cross-linked network. When the chemical reaction and macroscopic attraction between the two reach a balance or end, a mutually fused and well-defined intermediate layer will be formed between the metal layer and the AP-SMR. The existence of the intermediate layer provides a good transition and connection between the metal layer and the AS-SMR layer with completely different properties, so that the overall performance of the entire TSMP/M heterointerface can be coupled and improved. **Figure 6 g)-i)** shows the microscopic cross-section of the part after being bent to 90°. Although a small amount of metal layer peeling off locally occurs, the overall connection is still maintained, which is crucial to the functional integrity of the TSMP/M heterointerface important. Furthermore, the activity of AP-SMR was again confirm. **Figure 6 j) -l)** show a closed cavity structure due to the presence of air bubbles during printing. In the cross-section after FIB treatment, we found its existence, and Ni was deposited on its inner wall. This means that the plating bath can reach the internal structure of AP-SMR through the cross-linked network, and the existence of the cavity is a chemical Reactions provide the necessary venue. Furthermore, in some thin-walled structures, we found that the metal layer itself also exhibits significant delamination, that is, the amount of Ni deposited between different layers is regularly different. **Figure 6 m)** shows one such cross-section, where the Ni fraction exhibits pronounced, regular stratification (**Figure 6 O)**). However, this phenomenon does not occur in the distribution of Pd (**Figure 6 n)**). In fact, DLP 3D printing is based on the principle of surface forming, that is, a certain thickness of photosensitive resin is irradiated by a UV mask irradiated on one side and polymerized. In its microscopic state, different irradiation depths or even uniform irradiation depths are not completely uniformly irradiated, so the density of the cross-linked network formed is not quite the same. In the part where the double bond cross-linking reaction does not fully occur, the molecular network is sparse, which can provide more space for Ni deposition. However, this defect is expected to be exploited to realize the construction of functional interfaces with special designs by controlling the degree of crosslinking of polymer networks. Indeed, the Pd (0) embedded in the cross-linked network (after the reduction process) forms nanoscale anchors in the polymer. The electrochemical force (Van der Waals force, electrostatic force, chemical reaction force) [35,36] in ELP is the key factor to construct this kind of micro-nano interface. The plating solution can penetrate into the cavity formed by the cross-linked network to deposit a metal layer with better adhesion on the nano-anchor points. The existence of the intermediate layer effectively connects the non-metal layer and the pure metal layer and exists as a polymer-metal composite structure. This allows the TSMP/M heterointerface to simultaneously possess the properties and functions of metals, polymer-metal composite structures and polymers.

## 4D electronic structures with topology of 3D TSMP/M heterointerface

The TSMP/M heterointerface with an intermediate layer has good mechanical and electronic properties, and can be programmed in a specific pattern as a whole topology to build different functional structures, and thus can be fabricated as a 4D electronic smart structure by MM-DLP 3DP. 4D electronics will combine information processing capabilities (metal layer) and smart polymer functions (substrate layer), and the key to them as a whole is the intermediate layer. Such process advantages are beyond the reach of the existing pure polymer 4D printing process (Electronic features are missing) or pure metal 4D printing (Defects in processability).

The TSMP/M heterointerface can be integrally programmed in a hollow structure with a lattice structure so that the original part becomes a metal electrode structure (shown in **Fig. 7 a)** that can be controlled electrically as a stretchable scaffold (**I)-II)**). The special lattice design can make the part deform according to the designed shape, and change some original properties (shape, density and even size, **Video 5, Video 6**). In the high-density lattice design, the parts marked by TSMP/M heterointerface can also undergo electronically controlled shape programming and recovery as a metal foam (**III)-V), Video 7, Video 8**). The minimum manufacturing resolution of this type of structure can reach 40 μm, so it is expected to play a role in medical treatment as a cell scaffold. It is worth noting that the metal layer is not limited to Ni alone, Cu **(IV))**, Au **(V))** and other metals that can undergo metal deposition reactions catalyzed by Pd are all valid candidates for our technology [36]. Even, we can use the deformable features of the 4D polymer structure to build Ni, Cu, and Au three metals that exist simultaneously on the same plastic substrate as shown in **Fig. 7 a) VI)**. This will greatly broaden the cutting-edge applications of this technology in electronic heat dissipation, smart electronics and semiconductor industries.

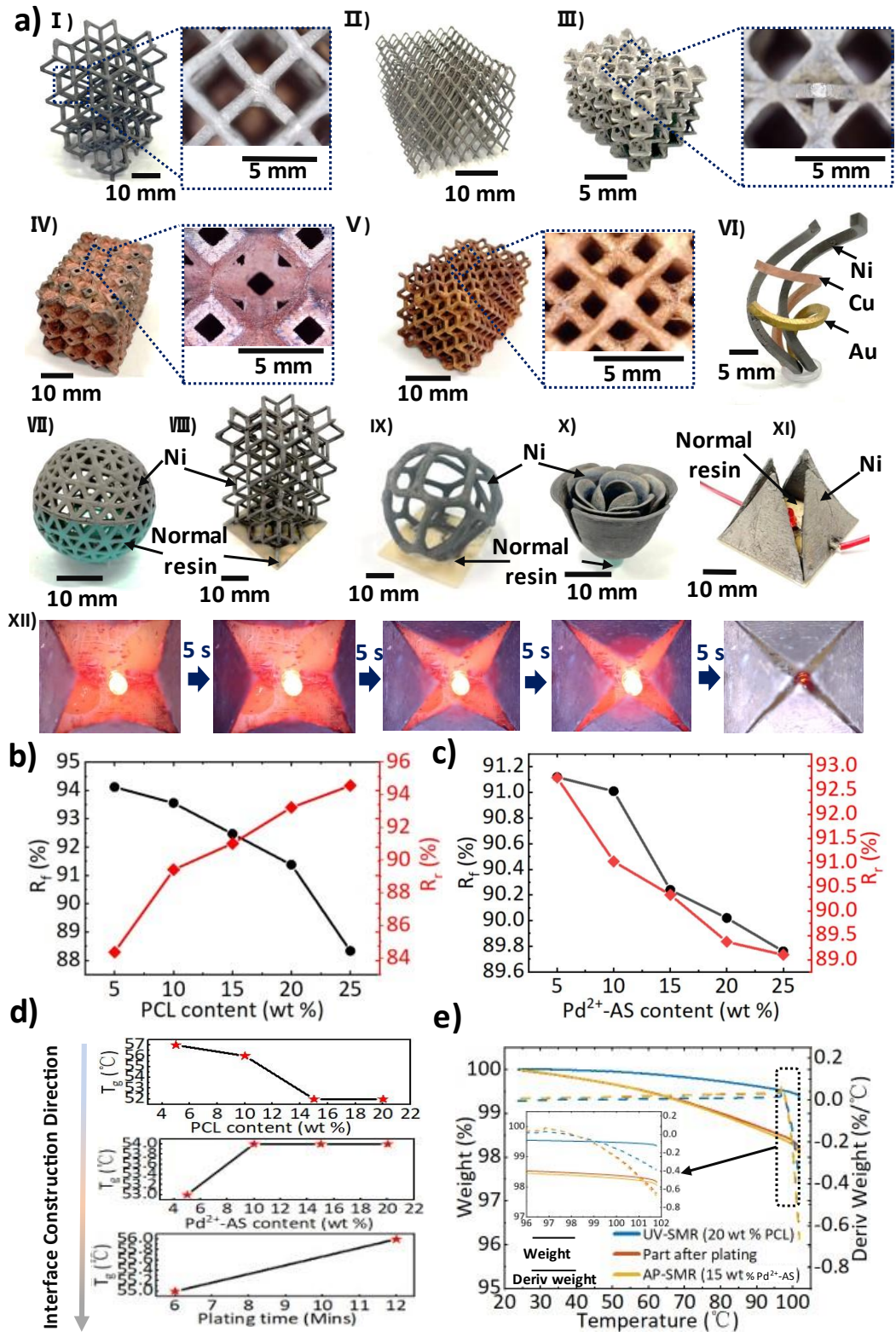

**Figure 7. 4D electronic structures with topology of 3D TSMP/M heterointerface and their comprehensive properties**. a) 4D electronics with different complex shapes and 3D programming patterns of the TSMP/M heterointerface, in which Ⅰ)-Ⅴ) fully programmed metamaterial parts; Ⅵ) plastic-metal composite structures with three different metal topologies, obtained by utilizing the deformable feature of AP-SMR; Ⅶ)-Ⅹ) 4D electronic structure constructed by 3D programming of

the TSMP/M heterointerface on normal resin; XI) is a 4D electronic pyramid structure, consisting of an LED circuit built on normal resin and four extendable shells; The process of the shell gradually closing in the extended state is shown in XII). When the closing is completed after 20 seconds, the LED is extinguished; b) The influence curve of PCL content on the $R_f$ and $R_r$ of UV-SMR; c) The influence curve of $Pd^{2+}$-As content on the $R_f$ and $R_r$ of AP-SMR; d) Tg temperature of UV-SMR, AP-SMR and the TSMP/M heterointerface after plating with different components; e) TGA and DTG thermograms for process samples of UV-SMR with 20 wt% PCL, AP-SMR with 15 wt% and part after plating.

The concept of 4D electronics can be more embodied in multi-material parts. Through MM-DLP 3DP, we can construct the patterned 3D TSMP/M heterointerface shape on any photosensitive functional polymer, and endow 3D electronics with intelligent properties. **Figure 7 a) VII)** shows a hollowed-out ball structure, in which half is a common polymer and the other half is programmed as a TSMP/M heterointerface. And the lattice structure with TSMP/M heterointerface (**I)**) can also be built on plastic parts to achieve selective deformation (compression and release) of the designed part of the structure, in which plastic parts can be used as frames perform certain tasks (VII)-IX)). Unlike the application of deformation processes, some structures designed to return to their original shape precisely to function, can also be constructed and realized. In the field of electronics, some structures with TSMP/M heterointerface are designed as electronic devices, such as the 4D electronic parts shown in **Fig.7 a) X)-XII)**.

As the driving unit of the smart structure, the TSMP/M heterointerface exhibits positive shape memory performance and thermal performance, which is critical as an actuator. Among them, shape fixity ratio ($R_f$) and shape recovery ratio ($R_r$) are two important parameters reflecting shape memory properties. The $R_f$ can indicate the ability of the SMPs to fix the mechanical deformation applied in the programming of process, while the $R_r$ demonstrates the ability of the material to recover to its original shape. They can be determined from **Equal. S1)**, **S2)**, respectively. **Figure 7 b)** shows the $R_f$ and $R_r$ curves of UV-SMR with different concentrations of PCL. As the substrate layer providing shape memory properties, with the increase of PCL content, its $R_f$ is lower, while $R_r$ is higher. This is consistent with theory, as PCL plays an important role as a core driver in the material with respect to shape memory properties. In AP-SMR, the addition of $Pd^{2+}$-AS does not significantly change its shape memory performance, and the slight effect is also due to its influence on the actual concentration of PCL and ABS-Like polymers (**Figure 7 c)**). Therefore, the $R_f$ and $R_r$ of the proposed TSMP/M heterointerface are kept within the positive interval of 90 %-95 %. Since the intelligence of the TSMP/M heterointerface is based on thermally actuated shape memory properties, the glass transition temperature is an important interface between actuation and implementation. Through the DSC measurement, we obtained the Tg of each state of the TSMP/M heterointerface along its construction direction, as shown in **Figure 7 d)**. In UV-SMR, as the concentration of PCL increased, the Tg of UV-SMR decreased from 57 °C at 5 wt% to 52 °C at wt15% and remained at this value. In the AP-SMR stage, the addition of $Pd^{2+}$-AP has negligible effect on Tg. For TSMP/M heterointerface, its Tg increases from 55°C at 6 Mins (metal layer thickness less than 1 μm) to 56°C at 12 Mins (metal layer thickness is about 2 μm), which indicates that the metal layer thickness leads to a slight upward trend in its Tg. Therefore, we set the Tg temperature of the TSMP/M heterointerface to 55°C when driving the fabricated smart structure. TGA was used to evaluate the thermal stability of all shape memory based element [37]. The representative TGA and DTG curves obtained from process samples at different stages are shown in **Fig. 7 e)**. The weight loss curves of UV-SMR, AP-SMR and the TSMP/M heterointerface are all obtained by one-step thermal degradation. Increase from room temperature to 600.00 °C at 10.00 °C/Min. The degradation of the memory-based materials has no significant degradation from 25-200 °C. The thermal degradation temperature is 300-500 °C, which is caused by the degradation of memory material. The findings suggested that three kinds of materials exhibit a positive thermal reliability in the operating temperature range. To further test the stability of the set operating temperature of the proposed material, increase from room temperature to 100.00 °C at 10.00 °C/Min, isothermal for 30 Mins. The weight loss percentage was within 5 % in the operating temperature range (deformation temperature: 20-60 °C) of the shape memory element, and the trend was the same for the three stages of the process samples, where the main reason for the weight loss was due to the volatilization of moisture, and the trend was consistent with the overall results.

The comprehensive performance of the TSMP/M heterointerface enables our technology to provide the necessary manufacturing technology with unique and innovative capabilities for intelligent devices with advanced concepts. The unique microstructure of metamaterials and the physical realization of material design are the most direct beneficiaries. This type of device usually has subversive manufacturing characteristics such as the manufacture of arbitrary complex and changeable geometric structures, personalization, and high-efficiency molding. However, there are few reports on the integrated manufacturing of non-metallic and metallic materials in existing research.

**Figure 8 a) I**) shows a folded structure of a 2D metamaterial fully programmed into the TSMP/M heterointerface. **Figure 8 a) V)-IX)** shows the process of its programmed shape (**Figure 8 a) IV)**) returning to the original shape under the control of 1.5 V voltage, which only takes 40 s. This process is usually completed by its deformation process certain tasks, such as the deployment of solar cells in space. The rapid and low-energy unfolding process depends on electrical conductivity of the surface so that the temperature of the surface can be raised rapidly (**Figure 8 a) II)**) and the uniformity of the formed electrothermal plant (**Figure 8 a) III)**).

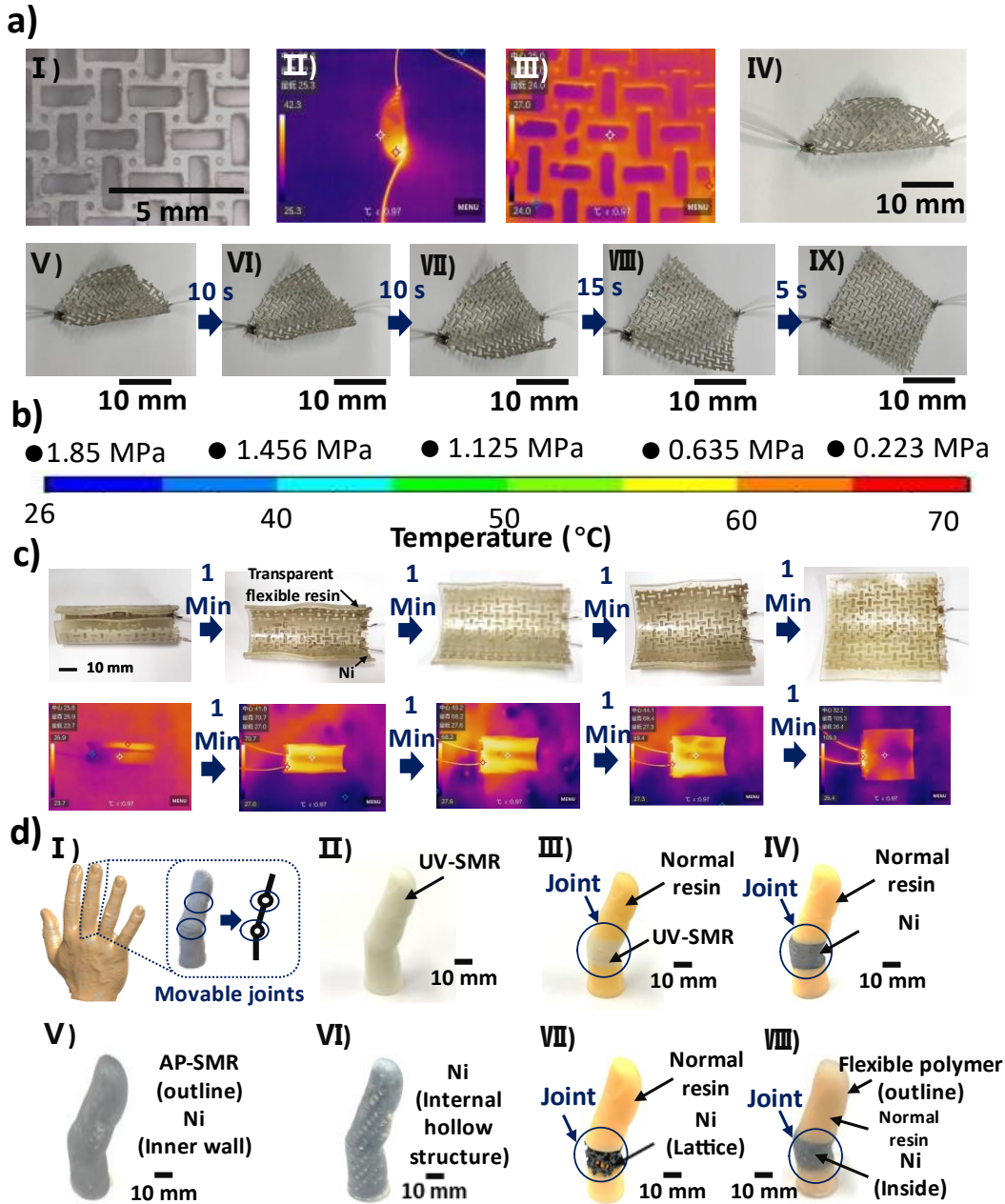

**Figure 8. 4D electronic structures with topology of 3D TSMP/M heterointerface for new concept**. a) The electronically controlled unfolding structure composed of 2D metamaterial structures is programmed as a TSMP/M heterointerface; b) The overall Young's modulus of the TSMP/M heterointerface changes with temperature; c) A bone-meat unfolded structure wrapped in 2D metamaterials with transparent flexible polymers; d) Finger robots implemented through different technical routes.

Furthermore, through MM-DLP3DP, we can pattern TSMP/M heterointerface on other functional materials (so-called 3D programming) to obtain more functions. **Figure 8 b)** shows the trend of the Young's modulus of the structure programmed by the TSMP/M heterointerface as the temperature changes, in which the Young's modulus of the structure decreases (softens) as the temperature increases. **Figure 8 c)** shows a bone-meat structural device, in which the bone structure is formed by the TSMP/M heterointerface in **Figure 8 a)** and the flesh structure is formed by a flexible polymer.

The 2D deformable structure embedded in the flexible polymer film can be unfolded and bent under electronic control to drive the flexible film to complete the process while protecting the driving structure from external damage. Such devices are expected to play an important role in the fields of flexible electronics and foldable electronics. Nevertheless, our technology is not limited to the additive creation of 2D smart devices, it is also handy for complex parts with 3D shapes. As shown in Figure 8 d), our technique can provide more flexible and efficient fabrication strategies by controlling the spatial distribution of heterogeneous materials and constructing ordered deformable structures.

Typically, 4D printing is used to support the potential of intelligent robots, while existing technologies exhibit certain limitations in terms of geometry, design, and material selection. The material combination capability of the proposed technology can provide more flexible, novel and intelligent solutions for robots. For the palm shown in **Figure 8 d) I)**, it can actually be simplified to a movable joint and a rigid rod structure. This makes it so that when fabricating a highly similar robot, its material division should consist of at least deformable smart materials and rigid materials. **Figure 8 d) II)** shows that the finger is made of pure UV-SMR, its shape is similar to that of a human finger, and it can have a shape memory effect (can bend but not like a human finger), and most technologies use this primary mode [38]. The emergence of multi-material 3D printing technology [39] enables the manufactured devices to further simulate the spatial division of different materials of fingers, as shown in **Figure 8 d) III)**. Since the drive still requires a thermal field, its deformation process cannot highly simulate a real finger. Furthermore, using the proposed technology, we can program specific structures with selective 3D interface, so that specific joints of fingers can be electrically controlled and driven selectively **(Figure 8 d) IV))**. Since the deformation of the solid structure is not easy, neither the shape programming nor the controllable deformation process of the finger is difficult. And our technology can provide support for a wider range of designs. The finger robot can be designed in the form of flesh and blood, the exterior is made of flexible materials and the interior is programmed by metallization (providing deformation actuation) through a thin-walled structures with identical contours as shown in **Figure 8 d) V)** or a hollow AP-SMR as shown in **Figure 8 d) VI)**. Alternatively, the joints of the fingers were fabricated as metal hollow shapes for more flexible deformation **(Fig. 8 d VII))**. Furthermore, the bone-meat structure model can be manufactured in the form of **Figure 8 d) VIII)**, so that the appearance is closer to the real finger in appearance, driving mode or the state of deformation. In conclusion, different forms of design have different advantages and disadvantages. As shown in **Table 1**, our technology can meet different purposes of design to support the designed parts to be more intelligent.

**Table 1 The shortcomings of different forms of finger-like robots can be manufactured by the proposed manufacturing technology**

| Robot form | Figure X c) | Similar in shape? | Selective Shape Programming | Selective drive | Similarity of deformation process | Available for other technologies? |
|---|---|---|---|---|---|---|
| Form-1 | II) | ∨ | ✗ | ✗ | ✗ | ∨ |
| Form-2 | III) | ∨ | ∨ | ✗ | ✗ | ∨ |
| Form-3 | IV) | ∨ | ∨ | ∨ | ✗ | ✗ |
| Form-4 | V) | ∨ | ✗ | ✗ | ✗ | ✗ |
| Form-5 | VI) | ✗ | ✗ | ✗ | ∨ | ✗ |
| Form-6 | VII) | ∨ | ∨ | ∨ | ∨ | ✗ |
| Form-7 | VIII) | ∨ | ∨ | ∨ | ∨ | ✗ |

The existence of the metal layer allows our control tentacles to extend to any part of the device, even inside, which cannot be stimulated by the thermal field. Even in the deformable structure **(Figure S7)** inside the completely closed structure, we can still realize its actuation through electronic design. This capability allows us to fabricate micro-actuator structures with internals and even realize their remote control, selective control, precise control and even physically isolated control.

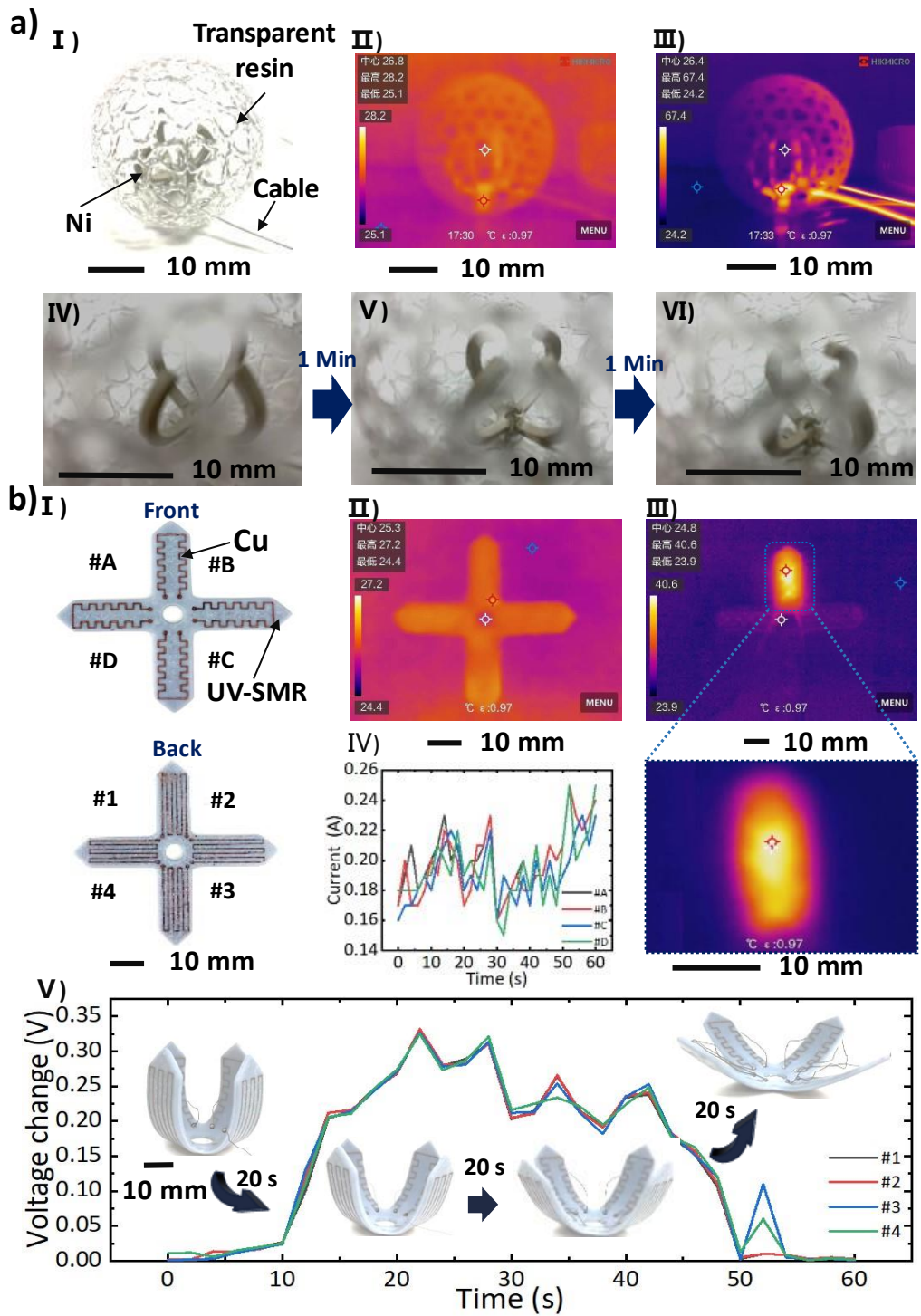

**Figure 9. A 4D electronic smart device with a new concept.** a) 4D electronic structure with micro actuators inside the part; by programming the AP-SMR mechanical gripper built inside the transparent hollow ball with the TSMP/M heterointerface, the internal micro actuator can be controlled offline; b) A 4D electronic smart device with a new concept. With double-layer TSMP/M heterointerface's 4D electronic intelligent mechanical claw integrates drive function and strain sensing function; among them, the front pattern can radiate a thermal field to drive the mechanical claw (UV-SMR substrate) to deform after being energized; in this process, the metal on the backside The pattern undergoes resistance changes during deformation to enable strain measurements.

**Figure 9 a)** shows the braking process of a mechanical claw machine built inside a hollowed-out ball under electronic control. The construction and shape programming of the TSMP/M heterointerface can be done through the voids in the external structure, while the external hollow structure acts as a barrier (**Figure 9 a) I)**). Through the electrical connection, we can precisely control the deformation of the mechanical claws in the hollow ball. **Figure 9 a) II)** shows the heat distribution image (infrared thermal imaging) after energization (0.8 V) for 1 s, showing a significant thermal field division. Under the control of the current, the mechanical claw completed the deformation process from the programmed shape to the original shape (**Figure 9 a) IV)-VI)**) in about 3 mins. **Figure 9 a) III)** shows the infrared heat of the part in about 3 Mins The molding map shows a more distinct temperature distribution, with the highest temperature reaching 67.4°C, while the outer hollow ball structure remains at room temperature. Further, the proposed technology is expected to be extended to any potential smart polymer materials as a platform technology, based on the kinetics of $Pd^{2+}$-induced metal deposition embedded in the polymer network. First of all, at the heterointerface level, the y-Metal heterointerface (y/Metal heterointerface) constructed by any smart polymer x has function A, which is analogous to the metal conductive interface in electronics; while MM-DLP3DP can realize y/Metal heterointerface on the functional substrate z (with function C) 3D selective programming (three-dimensional patterned construction), analogous to building a grouped circuit in a PCB board. In this way, the intelligence that couples x, metal (Ni, Cu, Au, etc.) and z is obtained Devices are realized macroscopically as including function A and B but surpassing their function C. **Figure 9 b)** shows one such paradigm, allowing the design of an electromechanical claw with in-situ displacement sensors. The front and back of the cross-shaped UV-SMR part are constructed with 8 TSMP/M heterointerfaces with different patterns (in this example, x=SMR, y=AP-SMR, z=SMR). Among the 8 Cu electrodes, the front four electrodes (numbered #A-#C respectively) are used for heating and radiating heat, while the 4 back electrodes (numbered # 1-# 4 respectively) are used for strain gauges to realize mechanical grippers The measurement of the amount of deformation **(Figure 9 b) I))**. **Figure 9 b) II)** shows the heat distribution map after the front electrode is energized (1.2 V) for 5 s, showing the uniformity of the heat distribution of the heating electrode (Mechanical claw driven by patterned thermoelectric stages seen **Video 9**). **Figure 9 b) III)** shows the infrared thermal imaging spectrum of the front electrode after 10 s of power failure. Although the current is stopped, its thermal field still shows patterned radiation in a short time (lower part, enlarged picture), which indicates that the features of control in low energy situations. By tracking the current value of each electrode in real time, the good electrothermal properties of the metal layer in the interface are again confirmed. **Figure 9 b) V)** shows the process of the mechanical claw stretching from the closed state to the open state within 1 Min. By tracking the voltage change Figure 9 b) Ⅳ) of each back electrode, we can measure the deformation of each mechanical claw.

The programmed topology exhibits active intelligence, and through the combined design of 3D patterns of the TSMP/M heterointerface and base materials, this technology is expected to create multiple forms of devices with completely new concepts in an integrated manner. Furthermore, the metal layer can be given more functions in a rich electrolytic plating process (electrolytic plating of pure Ni for magnetism) to obtain more experienced performance. This technology is expected to be extended to other cutting-edge smart materials as a platform technology to achieve unique performance for smart devices through the construction of heterogeneous interfaces.

## Electronically controlled 4D actuation device coupled in the smart system

A smart system is a complex system that integrates advanced electronics, sensors and control mechanisms. These systems are designed to mimic human intelligent behavior and decision-making processes for autonomous, adaptive, and cooperative task performance. Intelligent systems are generally able to sense the environment, collect and process data, and then make decisions and actions based on this data. The emergence of smart materials (4D materials) provides a novel material basis for smart systems, opening up a wide range of opportunities for the development and application of smart systems, which can improve the performance, reliability and functionality of the systems. **Figure 10** and **Figure 11** respectively show two new concept devices, 4D electronic chip and unmanned aerial robot (**UAV robot**), to prove the promotion effect of the proposed technology on intelligent systems. **Figure 10 a)** shows the design concept of the 4D electronic chip, 55 independently controllable grippers are integrated on the base plate of 20 cm × 40 cm. The designed part of the manipulator is designed with through holes, so as to realize the electrical link between the manipulator's executive part and the electrical interface. The lower part of the 4D electronic chip is the electrical pin that can be connected to the functional PCB (similar to the patch package of the chip), so it can be integrated in the peripherals (communication and current control) as a whole to realize the control of the manipulator array on it. The design incorporates arrays of micro-actuators into existing electronics to perform specific tasks. **Figure 10 b) I)** shows the actual image of the 4D electronic chip, which has a tiny scale, can complete the task alone or in cooperation at the microscopic scale. Small-scale devices integrated with micro-actuators have the potential to become 4D micro-electromechanical systems (**MEMS**). **Figure 10 b) II)** show the recovery process of a single mechanical gripper from the programmed shape to the original shape. Through the precise input of the power supply, a single mechanical claw is selectively energized (3.3 V) and triggers the shape memory effect within a short period of 30 s that be viewed in a thermal imaging perspective (**Figure 10 b) III)**). **Figure 10 c)** shows the effect of 3D manufacturing of parts. The small-scale and large-scale manufacturing capability broadens the application potential of this technology and provides a potential strategy for the integration of complex micro-actuators and advanced electronics.

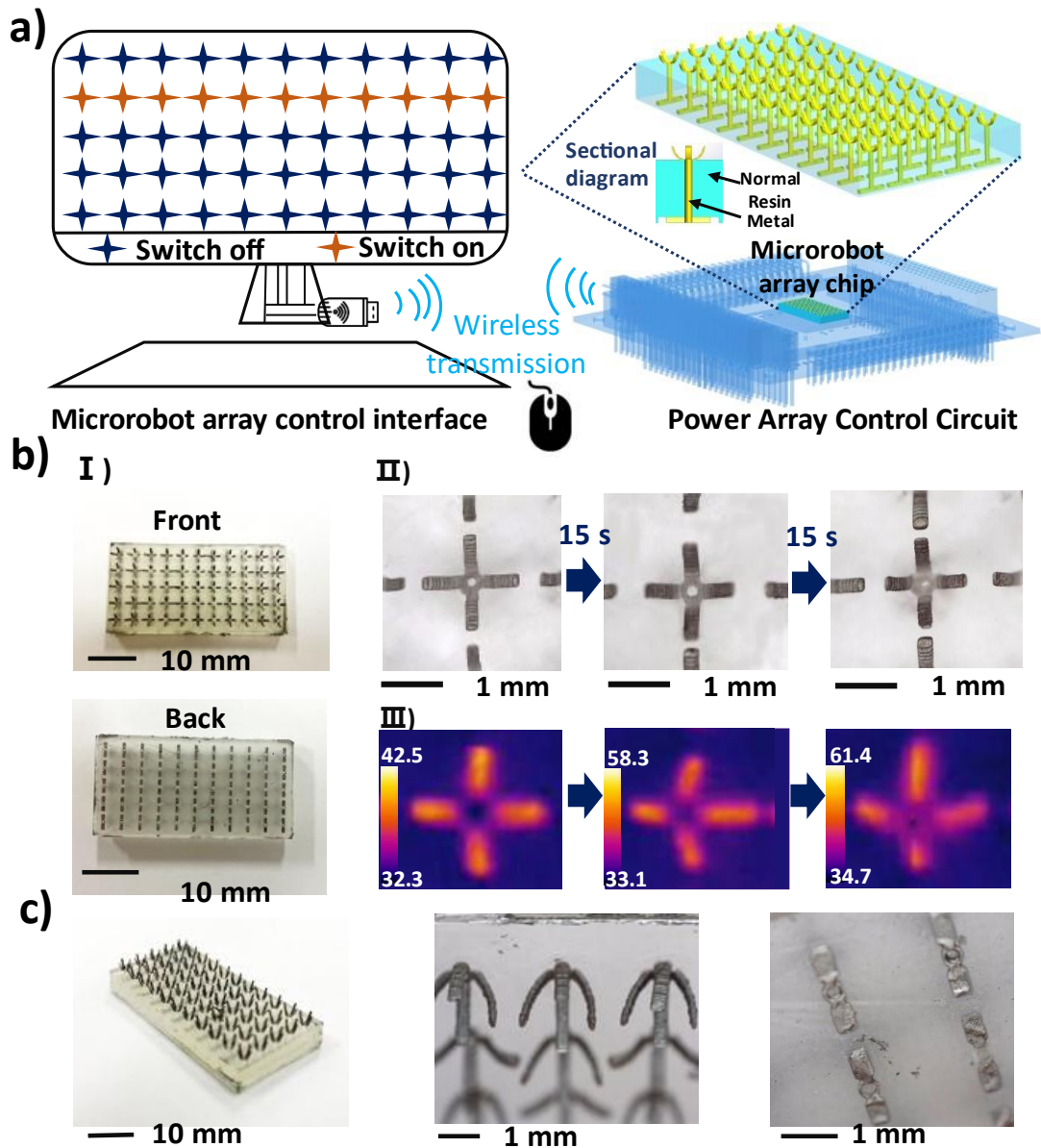

**Figure 10. A 4D system-on-chip integrating a micro actuator array on a chip.** a）Concept of a 4D chip that inherits the micromechanical claw array；b) The process in which a single mechanical claw is driven to deform; c）3D manufacturing effect of mechanical claw chip.

In **Fig. 11 a)**, a UAV robot system design is shown, which can complete the operation task on the basis of traditional UAV. The system consists of two robotic grippers with 4D electronics mounted on a flat-panel drone. The 4D electronics are integrated (integrated mechanical interface, electrical interface, and actuator as shown in **Fig. 11 b) I))** and are designed for target placement and grasping tasks. Compared with traditional mechanical claws, the 4D electronic form is lighter in weight and can be driven by micro battery (**Video 10**) or pulse-width modulation **(PWM) (Video 11)** waves in integrated circuit **(ICs)** therefore has better integration of intelligent systems. The preset tasks of the system can be described as the following three processes: First, the UAV takes off and flies towards the target area; when it reaches the drop target (Over base a*), the PWM wave drives #1 manipulator to open and place the red hollow ball on the target on the base a*; then, the UAV continues to fly towards the grab target to the position where the #2 mechanical claw is above the green hollow ball, and the #2 mechanical claw is driven by the PWM wave to complete the grabbing of the green ball; so far, the mission is completed and the drone flies away from the target area **(Video 12)**. **Figure 11 b) II)** show the real picture of the UAV robot and the image of the microcontroller unit **(MCU)** and battery and battery respectively, with 3D electronic traces, mechanical interfaces and actuators. **Figure 11 c) I)** shows the process of the UAV turning on the mechanical claw to release the ball (red) over base a*, which took about 72 s (from powering up #1 mechanical claw to disconnecting the drive signal after the ball was successfully dropped). **Figure 11 c) III)** shows the process of the UAV hovering over base b* and activating the mechanical claw #2 to grab the ball (green), which took a total of roughly 56 s. It is worth noting that all of the above processes are powered by a micro battery with an output of 3.7 V, which shows good controllability throughout the entire process thanks to the good integration with the intelligent system. **Figure 11 d)** respectively shows the thermal imaging images of activating the #1 gripper and simultaneously shutting down the #1 and #2 grippers with activating the #2 gripper, respectively. Although the UAV itself emits heat, it does not affect the control of the grippers.

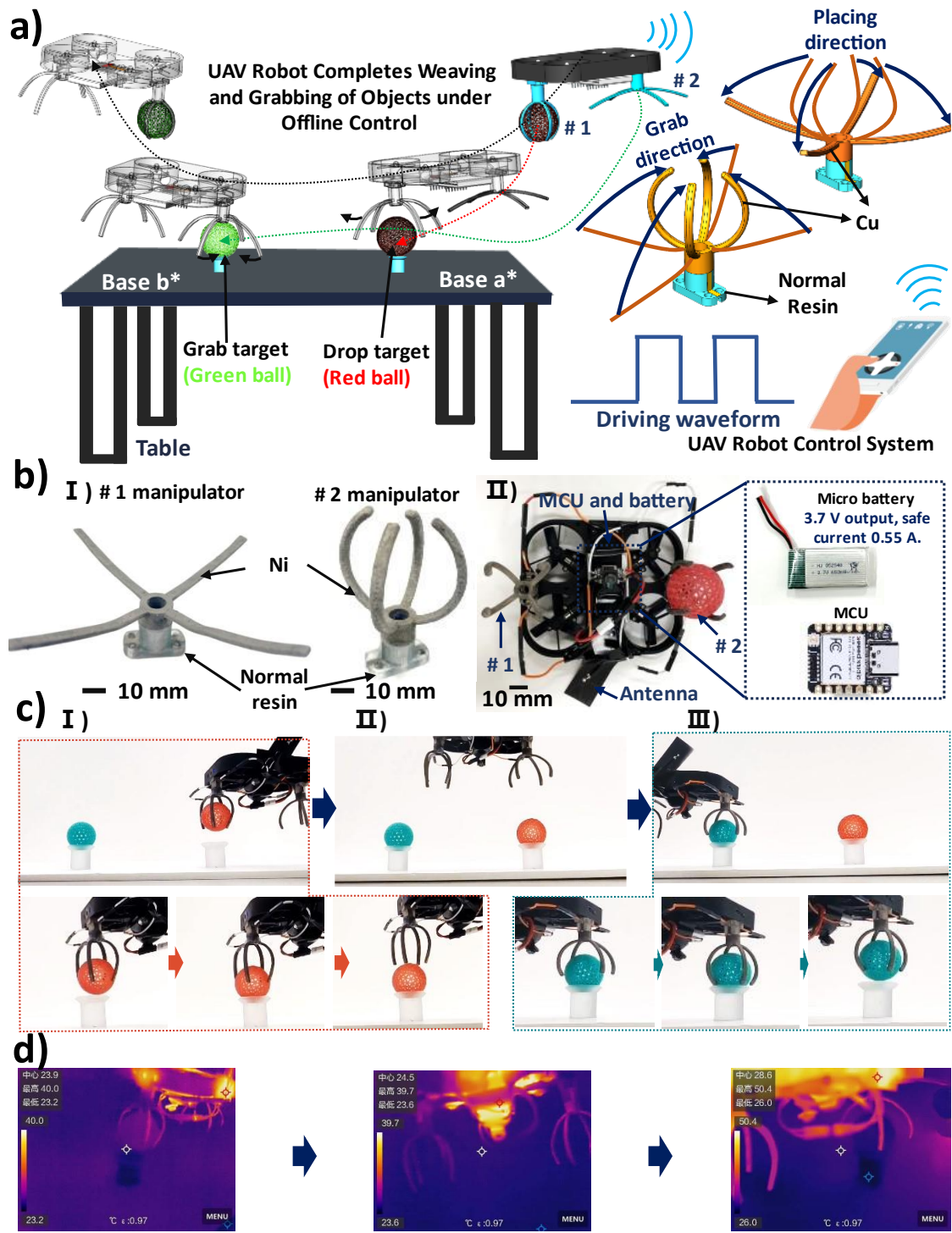

**Figure 11. The manipulator system integrated into the drone realizes an intelligent system to complete specific tasks.** a) The proposed 4D structure is fused with the concept of UAV robot system accomplishing a specific task paradigm; b) 4D mechanical claws of different designs can be directly installed with the UAV system as electronic devices and controlled by micro-battery drive and MCU; c) The process of UAV robot system placing and grabbing different target balls; d) The thermal imaging images of activating the #1 gripper and simultaneously shutting down the #1 and #2 grippers with activating the #2 gripper, respectively.

## Conclusion

We report a new smart device construction strategy by constructing a TSMP/M heterointerface with both smart polymer and metal properties and realizing its 3D programming on arbitrarily complex 3D functional parts through a multi-material DLP 3D printing process. Our novel 4D printing method relies on the combination of functional polymers and metals in the TSMP/M heterointerface. This technique has potential as a platform technology where functional polymer base materials can be extended to a wide range of functional materials (liquid crystal elastomers, ceramics, piezoelectric elastomer, dielectric elastomer, etc.) and metals (Ni, Co, Cu, Au, Ag, Pt, etc.). On this basis, the MM-DLP3DP further broadens the application potential of the proposed strategy. By constructing the TSMP/M heterointerface with the designed pattern on the 3D parts with specific functions, new smart devices with a new concept It is expected to be integrated and function. All in all, due to the intelligence and policy generalizability of our TSMP/M heterointerface, our research opens up new avenues for the frontier fields of informatics such as intelligent tissue engineering, 3D intelligent electronics, and intelligent robotics.

## *Methods and Experiments*

### *Materials*

The materials used in this study contain mainly pharmaceuticals required for the preparation of active precursors, various types of photosensitive resins and chemicals used to perform electroless plating. The chemicals were purchased from Japanese pharmaceutical companies, and the photosensitive resins were purchased from different companies according to the demand. For information on the materials used in the experiments, please see the ***supplementary material***.

### *Functional polymers formulations and solution preparation*

The solution configuration and functional resin preparation involved in this study mainly involves the preparation method of $Pd^{2+}$ based active solution, the preparation method of UV-SMR, the preparation method of AP-SMR, preparation of cleaning solution after printing. Wherein, the above solutions follow the chemical solution configuration guidelines while the desired solution is added to the functional resin and then mixed thoroughly and uniformly. Please refer to the ***supplementary material*** for detailed preparation methods. The preparation method of $Pd^{2+}$ based active solution, the preparation method of UV-SMR, the preparation method of AP-SMR, the preparation of cleaning solution after printing, please also see the ***Supplementary Material***.

### *MM-DLP3DP to achieve selective 3D programming of TSMP/M heterointerface*

An MM-DLP3DP equipment with three stations designed by our research lab is used to manufacture multi-material parts (**Figure S3**). In this configuration, the printer can choose between three pools (material pool A, material pool B, and cleaning boxes), and the printing platform can be moved in the X and Z dimensions to switch between them. The photosensitive material1, the active precursor, and the cleaning solution are held in material pools A, B, and the cleaning box 1-4, respectively. The device employs black-and-white liquid crystal display **(LCD)** technology to generate a transmissive graphics mask with built-in ultraviolet **(UV)** light at 405 nm, resulting in a light field associated with a specific slice pattern, allowing for layered curing and molding of parts. The entire 3D printed part moves as a unified unit, allowing for material changeover and ensuring exact interposition relationships between different material topologies inside the same part. Our previous study [25] describes the modeling, slicing, processing, and detailed printing procedures for the multi-material 3D printed model. Please see ***Supplementary Materials*** for further information on multi-material 3D printing device architecture (**Figure S3 b)**), procedures (**Figure S4**), printing parameters (**Table S2**) and some of the multi-material parts manufactured (**Figure S5**).

### *Selective metallization based on electroless plating*

The cleaned and dried parts containing AP-SMR are chemically etched and chemically reduced and then immersed in an electroless plating solution. When the desired build metal is Ni, the pH is 9 and the temperature is maintained at 70°C (**Table S3**). In each printed multi-material part, the active precursor, in which $Pd^{2+}$ ions are uniformly dispersed, is distributed on the resin substrate in a specific three-dimensional topology. Chemical etching increases the exposure probability of $Pd^{2+}$, and chemical reduction reduces $Pd^{2+}$ to Pd. When the part is immersed in the bath, Pd acts as a catalytically active metal core, triggering an ELP reaction in a specific microscopic area, resulting in target Ni metal deposition. Because the Pd is embedded in the surface of the active precursor component, there is no spillover or deflection of the deposited coating due to catalyst migration during the ELP process. See Supplementary Materials for ELP formulations (**Equal.S2-S5**), proportions and other metals of Cu (**Table S4, Equal. S6-S7**) and Au (**Table S5, Equal.S8-S9**).

### *Characterization and Experimental Testing*

The characterization experiments involved in this study are FTIR, SEM, XRD, XPS, Fib, EDS, DSC, TGA, and mechanical testing. The above experiments were all measured according to the consensus method, in which FTIR was performed by FT/IR-4200 (Japan Spectroscopy) equipment; SEM and EDS were performed by JSM-5500 (Japan Electronics) equipment; XRD was performed by X-ray refraction device-Miniflex (Rigaku); XPS was performed by JPS-9010MX electron spectroscopic device; Fib was performed by JIB−4000 concentrated ion beam processing device (Japan Electronics); DSC was performed by DSC-8500 (PerkinElmer, Inc, USA) equipment; TGA was performed by TG-8120 thermogravimetric differential thermal analyzer (Rigaku); mechanical testing experiments were performed by MCT-2150 bench-top tensile and compression testing machine (A&D). Other experimental tests include the test of 3D printing accuracy, the measurement of coating thickness, the test of bending fatigue life, and the measurement of resistivity of resistive coating. Please refer to the supplementary material for specific measurement methods.

## Supporting Information

● The restoration of the Eiffel Tower model from its programmed shape to its original shape driven by a thermal field. The Eiffel Tower model made by pure UV-SMR was programmed to bend and returned to the original shape when the printing was completed under the drive of the thermal field. (Movie S1) (.MP4)

● The restoration of the thin-walled parts composed of UV-SMR from its programmed shape to its original shape driven by a thermal field. The printing effect of thin-walled parts composed of UV-SMR, which supports aggressive molding properties. (Movie S2) (.MP4)

● The restoration of the microstructure parts composed of AP-SMR from its programmed shape to its original shape driven by a thermal field. The activated UV-SMR maintains the original shape memory characteristics and molding properties under 405 nm ultraviolet rays, while having the activity of inducing electroless plating. (Movie S3) (.MP4)

● The middle part of the Eiffel Tower is 3D programmed with TSMP/M heterointerface and its process of being selectively electronically controlled. The TSMP/M heterointerface can generate heat after being energized, so it can be selectively programmed in shape and return to its original shape under the control of a small voltage. (Movie S4) (.MP4)

● Electronically controlled deformation process of microscopic scaffold parts (low lattice density) with special crystal lattice. The scaffold is integrally programmed with TSMP/M heterointerface (Ni), which can be restored from the programmed shape to the original shape within minutes when driven by a 3.3 V DC power supply, demonstrating its potential as a cell scaffold. (Movie S5) (.MP4)

● Electronically controlled deformation process of microscopic scaffold parts (low lattice density) with parallelogram crystal lattice. The scaffold is integrally programmed with TSMP/M heterointerface (Ni), which can be restored from the programmed shape to the original shape within minutes when driven by a 3.3 V DC power supply, demonstrating its potential as a functional electrode. (Movie S6) (.MP4)

● The deformation process of metal foam (electronically controlled expansion) with higher lattice density programmed entirely as a TSMP/M heterointerface (Ni) in the microscopic state. Metamaterial structures with higher lattice densities can deform slightly and appear as metallic foams on a macroscopic scale. (Movie S7) (.MP4)

● The deformation process of metal foam (electronically controlled expansion) with higher lattice density programmed entirely as a TSMP/M heterointerface (Au) in the microscopic state. Metamaterial structures with higher lattice densities can deform slightly and appear as metallic foams on a macroscopic scale. (Movie S8) (.MP4)

● By 3D programming metal patterns (3D TSMP/M heterointerface (Ni)) on the surface of UV-SMR (or other smart materials), it can be deformed from the programmed shape to the original shape driven by a hot electrode (3.3 V DC power supply). It indicates the potential rich material combination capabilities provided by this technology. (Movie S9) (.MP4)

● The TSMP/M heterointerface has good electrothermal properties (characteristics of its metal layer), so the 4D mechanical gripper programmed by it can be directly driven by the micro battery. It indicates the application potential of the proposed technology in the electronics industry. (Movie S10) (.MP4)

● The selectively programmed Eiffel Tower model realizes the shape memory effect driven by the GPIO of the microprocessor. The entire system is driven by a 3.3 V micro-battery, and the PWM drive components with a GPIO output frequency of 1 HZ complete the transformation. (Movie S11) (.MP4)

● The manipulator system integrated into the drone realizes an intelligent system to complete specific tasks. Two 4D electronic mechanical claws of different designs are installed on the drone and cooperate with them to complete the placement and grabbing of the small ball. (Movie S12) (.MP4)

● Information on the materials used in the experiments; Original mechanical properties of the ABS-like photopolymer used; The preparation method of $Pd^{2+}$ based active solution; The preparation method of UV-SMR; The preparation method of AP-SMR; The preparation of cleaning solution after printing; Stability of the prepared PCL solution, UV-SMR and AP-SMR; Polymer flow deformation (viscosity) analysis; Measurement of Young's modulus; Multi-material 3D printing; 3D printing parameters set in this study; Measurement methods and experiments of 3D printing parts accuracy; Some multi-material parts manufactured by MM-DLP3DP (Before electroless plating); Mechanical properties of resin and its corresponding active precursor; Preparation of plating bath; 3D selective electroless plating; 4D structures in closed environments; Some supporting characterizations and descriptions of experimental methods. (.Word)

**Acknowledgments:**

The authors thank Kaoru KAWARABAYASHI Researcher, Noriko HANZAWA Researcher and Kagami Memorial Research Institute for Materials Science and Technology, Waseda University for helping to experiments. Thank Mikiko SAITO and Nano-Life Innovation Research Organization, Waseda University for the support of the chemical experiment environment. Thank Advanced Research Infrastructure for Materials and Nanotechnology (ARIM). Thank Ms Koh Joo Luan, Assistant Manager, Nanyang Technological University for experiments support. Also thank Yifan Pan from Waseda University for her help in carrying out the experiment. **Funding**: This work was supported by Singapore Ministry of Education [RG140/20], NTUitive Pte Ltd [NGF-2022-11-020], JST SPRING, Grant Number JPMJSP2128, JST-Mirai Program Grant Number JPMJMI21I1, and Kakenhi Grant Number 19H02117 and 20K20986, Japan, and Frontier of Embodiment Informatics: ICT and Robotics, under Waseda University's Waseda Goes Global Plan, as part of The Japanese Ministry of Education, Culture, Sports, Science and Technology (MEXT)'s Top Global University Project.

**Credit authorship contribution statement:**

**Kewei SONG:** Methodology, Investigation, Writing - original draft, Writing - review & editing, Visualization, Data curation. **Chunfeng XIONG:** Visualization, Data curation. **Ze ZHANG:** Visualization. **Kunlin WU:** Data curation. **Weiyang WAN:** Data curation. **Yifan WANG:** Conceptualization, Methodology, Investigation, Visualization, Supervision, Project administration, Funding acquisition. **Shinjiro UMEZU:** Conceptualization, Methodology, Investigation, Writing - review & editing, Visualization, Supervision, Project administration, Funding acquisition. **Hirotaka SATO:** Conceptualization, Methodology, Investigation, Writing - review & editing, Visualization, Supervision, Project administration, Funding acquisition.

**Competing Interest:**

The authors report no declarations of interest.

**Data and materials availability:**

All data needed to evaluate the conclusions in the paper are present in the paper and/or the Supplementary Materials. Additional data related to this paper may be requested from the authors.